\documentclass[conference]{IEEEtran}
\IEEEoverridecommandlockouts
\usepackage{cite}
\usepackage{amsmath,amssymb,amsfonts}
\usepackage{algorithmic}
\usepackage{graphicx}
\usepackage{textcomp}
\usepackage{xcolor}
\usepackage{CJK}
\usepackage{verbatim}
\usepackage{multirow}
\usepackage{subfigure}
\usepackage{color}
\usepackage{array}
\usepackage{marvosym}
\usepackage{balance}
\newcolumntype{C}[1]{>{\centering}p{#1}}
\def\BibTeX{{\rm B\kern-.05em{\sc i\kern-.025em b}\kern-.08em
    T\kern-.1667em\lower.7ex\hbox{E}\kern-.125emX}}

\usepackage{geometry}
\begin{document}
\newgeometry{left=19.1mm,right=19.1mm,top=24mm,bottom=19.1mm}

\title{Collaborative Visual Inertial SLAM for Multiple Smart Phones}

\author{\IEEEauthorblockN{Jialing Liu\textsuperscript{1},
		Ruyu Liu\textsuperscript{1}, Kaiqi Chen\textsuperscript{1},
		Jianhua Zhang\textsuperscript{2,\Letter}\thanks{\textsuperscript{\Letter}Jianhua Zhang is the corresponding author, zjh@ieee.org. }, 
		Dongyan Guo\textsuperscript{1}}
	\IEEEauthorblockA{\textsuperscript{1}College of Computer Science and Technology,
		Zhejiang University of Technology\\ 
		Hangzhou, China\\
		\textsuperscript{2}School of Computer Science and Engineering,
		Tianjin University of Technology\\
		Tianjin, China\\
}
\thanks{This publication was partially funded by the National Natural Science Foundation of China (61876167, 62020106004, 92048301) and the Natural Science Foundation of Zhejiang Province (LY20F030017).}
}

\maketitle

\begin{abstract}
	The efficiency and accuracy of mapping are crucial in a large scene and long-term AR applications. Multi-agent cooperative SLAM is the precondition of multi-user AR interaction. The cooperation of multiple smart phones has the potential to improve efficiency and robustness of task completion and can complete tasks that a single agent cannot do. However, it depends on robust communication, efficient location detection, robust mapping, and efficient information sharing among agents. We propose a multi-intelligence collaborative monocular visual-inertial SLAM deployed on multiple ios mobile devices with a centralized architecture. Each agent can independently explore the environment, run a visual-inertial odometry module online, and then send all the measurement information to a central server with higher computing resources. The server manages all the information received, detects overlapping areas, merges and optimizes the map, and shares information with the agents when needed. We have verified the performance of the system in public datasets and real environments. The accuracy of mapping and fusion of the proposed system is comparable to VINS-Mono which requires higher computing resources.
\end{abstract}


\section{Introduction}
With the popularization of smart mobile terminals, augmented reality (AR) industrial application scenarios are becoming more and more abundant. The future trend of AR is definitely towards a durable, multi-user, and shareable AR experience with greater immersion. AR technology can be abstracted into calculating the position of the camera and the 3D structure information of the environment through the image and other sensor information. The core of realizing multi-user sharing AR space is multi-intelligent mobile terminal collaboration simultaneous localization and mapping (SLAM), so far there is no multi-intelligent mobile terminal collaboration SLAM. Although there is already some SLAM for multi-agent collaboration, it has higher requirements for devices and is not suitable for the mobile terminal. 


Some of the work focuses on the single agent SLAM system, while some studies the self-localization between agents or the fusion of information from multiple agents to build a unique, global map. while few people study the multi-agent collaborative SLAM which can coordinate to complete the exploration task, make full use of the experience gained by other agents in one task, reduce the exploration time, improve the robustness of SLAM mapping, and obtain more environmental information in a short time. However, the multi-agent system exploration task also brings many challenges, such as local submap fusion, global map optimization, network bandwidth, and information reuse.

In this paper, we propose a centralized monocular visual-inertial SLAM that supports cross-device, multi-intelligence mobile terminal collaboration. Although currently it is designed for multiple ios devices, such as the iPhone and iPad, it can be easily extended to other smart phones. Each mobile terminal runs a visual-inertial odometry (VIO) module independently, and the measurement information is shared with the server through the network. The server manages all the information received from the client, detects the public areas, and then merges the map information constructed by multiple mobile terminals into one map. At the same time, it uses the measurement information of other mobile terminals to optimize the map constructed by another mobile terminal and informs the agent. The centralized architecture is chosen because the server has more computing resources and lower requirements for the computing power of mobile terminals. Therefore, for mobile terminals with limited computing power, any tasks with expensive computing requirements and no real-time requirements can be outsourced to the server. At the same time, it is ensured that tasks that are vital to agent autonomy are still running online, such as VIO.

The main contribution of the SLAM framework presented in this paper is as follows:
\begin{itemize}
\item A collaborative monocular visual-inertial SLAM system for smart phones. As far as we know, this is the first multi-agent collaboration SLAM system running on the mobile phone and supporting cross-device collaboration. 

\item The collaborative SLAM on mobile has achieved comparable performance with VINS-Mono\cite{qin2018vins} running on PC, such as single-agent mapping performance and multiple agent fusion performances.

\item An accurate and robust fusion strategy between multiple maps, where a local window is constructed for bidirec-\restoregeometry tional  reprojection error optimization, and consequently a more accurate transformation matrix between the two frames can be obtained. 

\end{itemize}

\begin{figure*}[htbp]
	\begin{center}
		\includegraphics[scale=0.065]{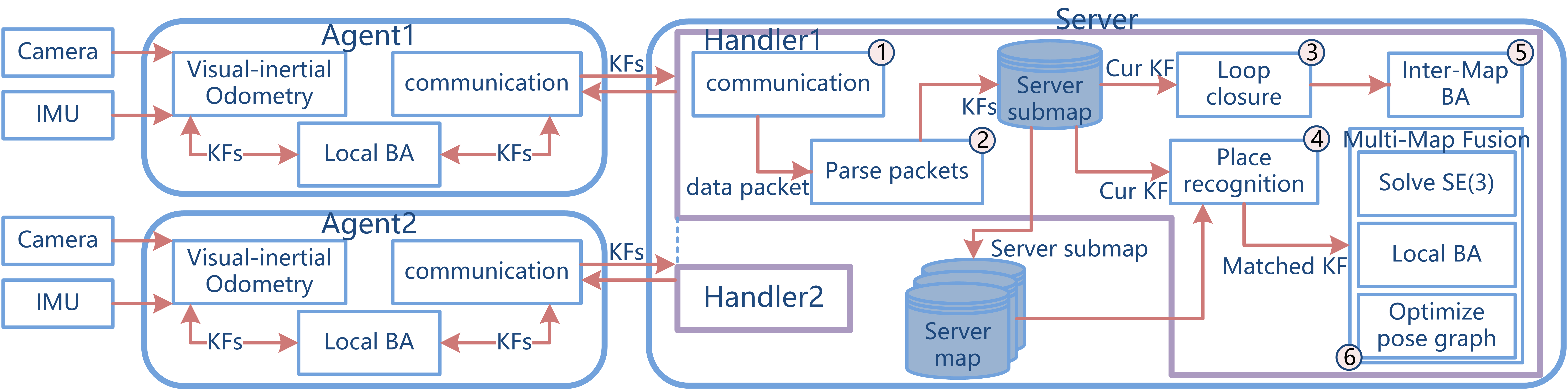}
	\end{center}
	\caption{ The proposed system components and data exchange. The mobile agent (e.g., iPhone 6S) runs a real-time VIO module, and a communication module to exchange data with the server. The server is a computer with higher computing resources, which performs non-real-time requirements and vital tasks: server mapping, place recognition, inter-map BA, multi-map fusion. BA: bundle adjustment; KFs: keyframes; MPs: map points; SE(3): Lie group.  }
	\label{fig.1.}
\end{figure*}

\section{RELATED WORK}
The researching work on state estimation based on VIO is quite extensive. The classical tightly coupled algorithms include MSCKF \cite{mourikis2007multi} and ROVIO \cite{bloesch2015robust}. Weiss proposed classical filter-based loosely coupled algorithm SSF \cite{weiss2012versatile} and MSF \cite{lynen2013robust}. Sibley\cite{falquez2016inertial} added the pose transformation generated by visual odometry (VO) calculation in the optimization framework of the inertial measurement unit (IMU), but the effect is not as good as tight coupling. OKVIS \cite{leutenegger2015keyframe} and VINS-Mono are the classic ones in tight coupling. The robustness of pose estimation and the effective management of maps have enabled the VIO field to develop very well. This marks the maturity of VIO in a single agent and at the same time paves the way for the development of multi-agent SLAM.

In the field of multi-agent collaboration SLAM, two problems are mainly solved: localization and map fusion. A few works in the literature deal with localization, the purpose of which is to estimate the relative positions of agents, or to locate the position of an agent in an existing map. The place recognition algorithm used by many recent visual SLAM and VO systems is based on bag-of-words visual place recognition. This method was proposed by Sivic \cite{sivic2003video}, and then improved by Nister \cite{nister2006scalable}, G{\'a}lvez-L{\'o}pez\cite{galvez2012bags} and DLoopDetector\footnote{DLoopDetector comes from https://github.com/dorian3d/DLoopDetector}. Ye  \textit{et al.}\cite{ye2017place} compared place recognition in semi-dense maps using geometric and learning-based approaches.

In the aspect of map fusion, Bonanni \textit{et al.} \cite{bonanni20173} proposed the fusion of 3D maps based on pose graph. Compared with traditional map fusion using a rigid body transformation, the map is modeled as a deformable map, and regarding other maps as observations, which can effectively deal with common deformation problems in the mapping to achieve higher accuracy. In our recently work, a fast matching strategy among multiple submaps is proposed for map fusion\cite{zhang2020map}. ORB-SLAM3 \cite{campos2020orb} discards the imprecise camera positions to reduce the accumulated error. Under the condition of ensuring the accuracy of the map, a rigid body transformation is used to seamlessly integrate the map. Among them, the ORBSLAM-Atlas\cite{elvira2019orbSLAM} module uses camera pose error covariance to estimate the observability of the camera pose to determine whether to retain the camera pose and create a new map.

There are two frameworks for multi-agent collaborative SLAM: centralized SLAM and distributed SLAM. Schmuck\cite{schmuck2017multi} proposed a multi-UAV collaborative monocular SLAM, where each agent independently explores the environment, and VO information is sent to a central server with higher computing resources. The server manages submaps for all agents, and detecting overlapping areas triggers map fusion, optimization, and distribute of the information back to the agent. Subsequently, Schmuck\cite{2018CCM}  proposed CCM-SLAM, where each agent only retains the basic VO to ensure that the agent can independently build the map in the environment and send the estimated pose and 3D point cloud map of the environment to the server for processing. The agent only maintains a local map with the latest $N$ keyframes. In addition to basic place recognition, map fusion, and global optimization, the server also adds redundant detection and deletion and information reuse between multiple agents. In the same year, Karrer\cite{2018CVI} also proposed a CVI-SLAM, a collaborative visual-inertial SLAM, in which each agent outsources computationally intensive tasks to the server, such as global optimization that consumes expensive computing resources of agents. The server also adds a detection and deletion of redundant information. The centralized SLAM for mobile devices\cite{zhang2020map} is based on the VO module of ORB-SLAM and integrates IMU information. After detecting the overlapping area, it immediately performs map fusion, but it does not perform an effective global map fusion optimization, and tracking failure is more likely to occur when running in the real environment.

In multi-agent distributed collaboration SLAM, agents share information and map online without relying on the server. Cieslewski \textit{et al.} \cite{cieslewski2018data} propose a DSLAM system that sends a compact and complete image descriptor calculated by NetVLAD to only one agent. Subsequently, Cieslewski and Scaramuzza \cite{cieslewski2017efficient} proposed a decentralized  visual place recognition system for networked agents. Wang \textit{et al.} \cite{wang2019active} proposed an active rendezvous for multi-agent pose graph optimization using sensing over Wi-Fi, which realizes on-demand information exchange through active rendezvous without using the map or agent location.


Compared with these collaborative SLAM systems, our proposed system adds a communication module from server to agent. Instead of fusing maps with a rigid transformation, all maps are modeled as deformable maps, and the fused maps optimize each other. This solves the common map distortion problem. Moreover, with the increase of overlapping regions among several submaps, more fusion optimization will be carried out among multiple submaps and the better integration can be achieved.

\section{SYSTEM OVERVIEW}
The proposed centralized SLAM system supports cross-device for multiple intelligent mobile terminals and a central server, whose architecture is illustrated in Fig. \ref{fig.1.}. The proposed system is designed to offload tasks that require high computational resource but not real-time performance to the server while ensuring that the autonomous modules of the agent run online on the agent. Each agent runs a real-time VIO module to estimate its poses and 3D points of the surrounding environment. In the local BA module, we define a sliding window consisting of $N$ frames near the current frame. In this sliding window, all frames should be optimized which requires high real-time performance. Meanwhile, this module can also ensure the accuracy of the agent. Even if the agent is disconnected from the server, it can also optimize the pose in the short term. VIO can put the measured value of each frame into the sliding window for optimization, limit the number of poses to be optimized in the sliding window, and constantly move out some frames, to prevent the number of poses and features from increasing with time, so that the optimization problem is always within a limited complexity and will not increase with time.

The server creates a server submap for each agent, which stores all the information sent by the agent, and all the server submaps corresponding to agents are put into a global server map container. Each agent submap uses a world coordinate system. The proposed system does not assume any prior information, or the configuration considers the initial position between any two agents. All agents operate independently until the place recognition module detects that there is a common area between the two submaps and therefore the server associates the corresponding agent's measurements. Each server submap executes a place recognition module. If two matched frames belong to the same submap and the map is not fused with other maps, the inter-map BA module will be performed, and the corresponding agent will be notified of the drift of the pose, which ensures the consistency between the agent-side and the server-side map. Otherwise, the multi-map fusion module optimizes the current agent's map using experience gained by other agents in the task.

\section{SYSTEM MODULES}
The various functional modules of the proposed system shown in Fig. \ref{fig.1.} are described in detail below.
\subsection{KF-based VIO and mapping}


There is no doubt that every VIO system can be used as a front end for agents to process new frames, as long as it is keyframe-based. The VIO module used in the proposed system is the VIO module of Vins-Mobile \cite{li2017monocular}, because it is one of the best monocular VIO available at that time, and it can run stably on the ios system. The communication module is integrated into this VIO system. To be able to experiment on public datasets, such as EuRoc\cite{2016The}, we also add specific processing of the data, such as dedistortion, reading and flipping of image data, and alignment of measurement units.

The server map container contains all the server submaps created by the agent handler and contains all the experience gained by the agent in one task. After initializing the system, each time an agent requests to connect to the server, an agent handler is created for the agent. When at least two agent handlers are created, the server creates a server map container. A multi-map fusion optimization is carried out when the server detects the common area of two submaps, and the two submaps are fused into one map.

\subsection{Agent handler}
The agent handler manages the data that the agent sends to the server and forwards messages to the server when necessary. As shown in Fig. \ref{fig.1.}, for each agent, an agent handler module is instantiated, creating six modules that run on five separate threads, where the fifth and sixth modules are running on the same thread. It creates a communication module to communicate with the agent and a module that parses data packets into keyframe information, initializes a server submap for each agent, puts it in the server map, and establishes a server submap manager. The submap manager can detect loop inside the submap, identify the common areas of multiple submaps, run global optimization within submap or between multiple submaps.

For the communication module, a keyframe has an average of 70 map points, 700 fast corners, and the corresponding number of 256-bit descriptors. Considering the large amount of data sent by network communication at one time, the efficiency of automatic packet cutting and packet splicing is lower than that of manual packet cutting and data splicing, and the transmission frequency and density also affect the communication efficiency. Besides, there is the phenomenon of packet loss in the actual situation, and the loss of a large packet means a larger amount of retransmitted data. Therefore, we divide the packet into 4*1024 sizes artificially and set the cache area for receiving data on the server to 210*1024. The data sent by the server to the agent is almost only the pose drift of the keyframe obtained by global optimization. The amount of data for this is very small, so the data packets communicated between the server and multiple agents are mainly the map information sent by the agent to the server. This is also proved in Fig. \ref{fig.4.} below.

\subsection{Place recognition}

The place recognition module detects the common area between a keyframe and the location in the server map container. Since each submap sent by the agent is incremental, the position overlap is detected each time with the latest frame. For each keyframe received by the server, two types of location recognition retrieval can be performed. One is place recognition with the internal map. When the place recognition module detects a pair of matching keyframes inside the agent ($KF_{old}$, $KF_{cur}$), if the map has been fused with other maps, it will trigger multi-map fusion module, otherwise, it will trigger inter-map BA module. The other is to match the location in the server map container. If the common area between agents is detected, a $SE(3)$ transformation between two matching keyframes is calculated, and more constraints are added between the two submaps, which is also the condition of the map fusion module.

\subsection{Visual multi-map merging}

\begin{figure}[htbp]
 	\centerline{\includegraphics[scale=0.075]{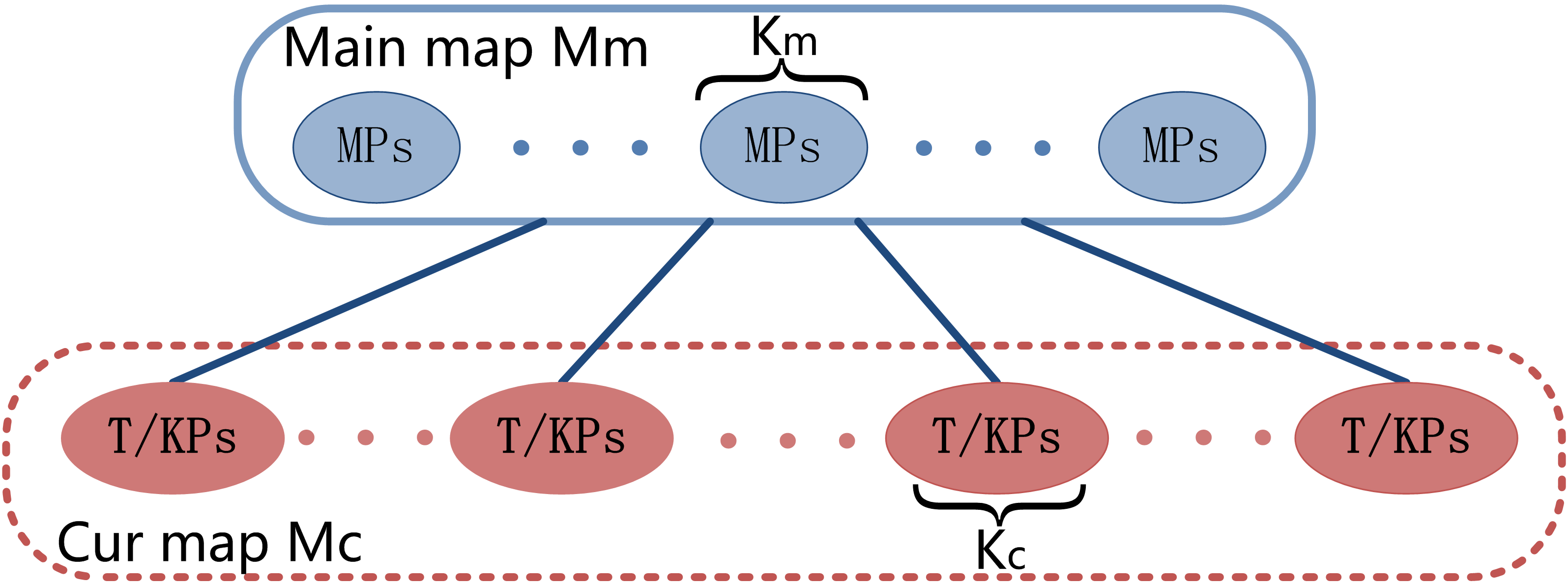}}
 	\caption{The cost function of local window optimization.  The blue ellipse represents the 3D map points of each frame belonging to map $M_ {m}$ in the local window. The red ellipse represents the 2D feature points of each frame of the map $M_{c}$ in the local window and the pose $T$ in the map $M_{c}$ world coordinate system. The blue solid line represents the reprojection relationship. MPs: 3D point clouds; T: Euclidean transformation; KPs: 2D keypoints; Km: matching frame of the map $M_{m}$; Kc: the search frame of the map $M_{c}$; }
 	
 	\label{fig.2.}
 \end{figure}

\begin{figure}[htbp]
 	\centerline{\includegraphics[scale=0.105]{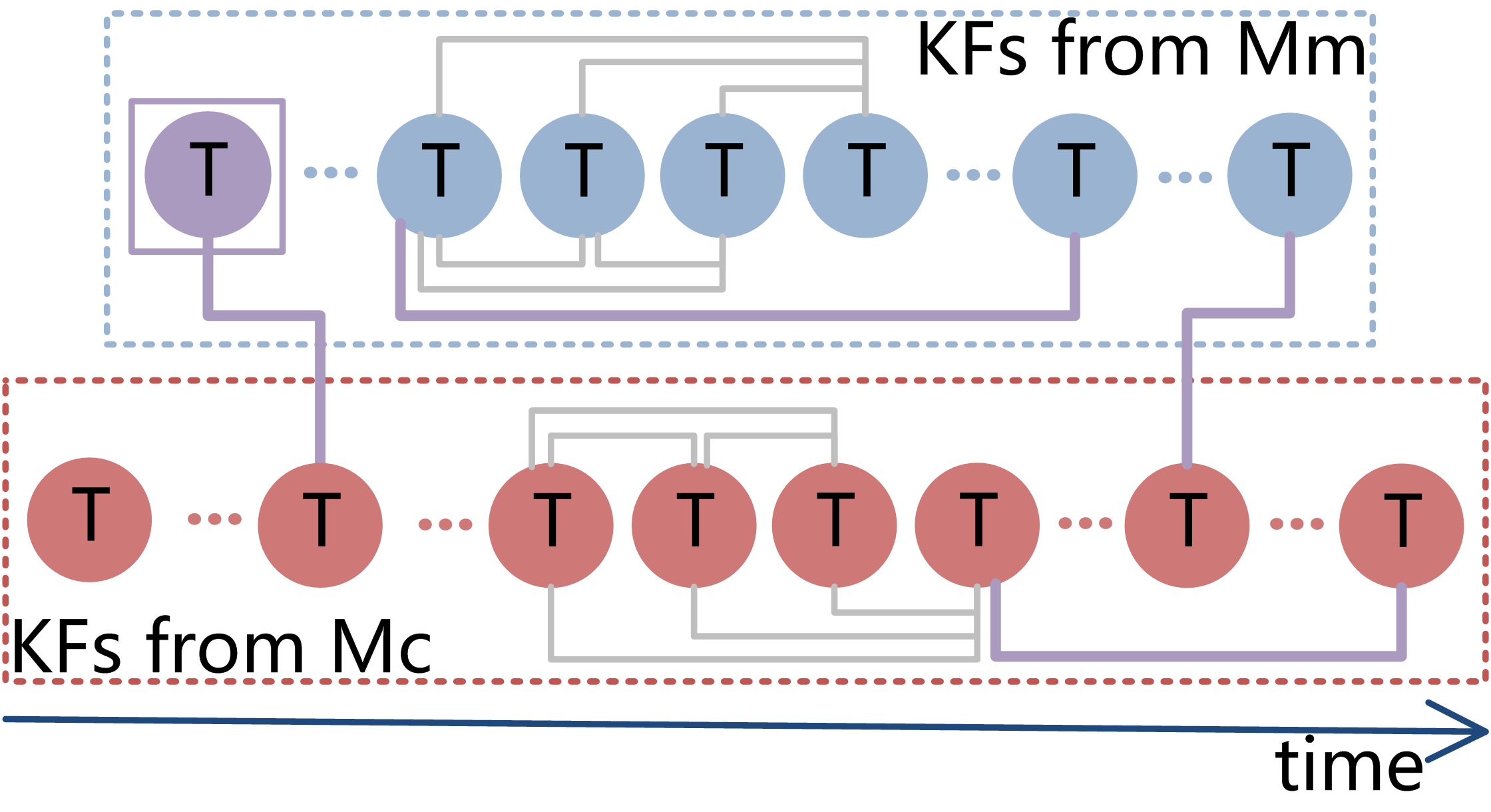}}
 	\caption{The indirectional cost function of multi-map fusion optimization. The blue box represents the map $M_{m}$, the red box represents the map $M_{c}$, and the circle represents the pose of the frame participating in the optimization. The purple circle represents the pose of the first loop frame in the map $M_{m}$ to participate in local window optimization. The purple box indicates the optimization variable fixed in it. The purple solid line between circles indicates that two frames are matched frames, and the gray solid line indicates the constructed sequence edge.}
 	\label{fig.3.}
 \end{figure}

When the place recognition module detects a pair of matching keyframes ($K_ {c}$, $K_ {m}$), which belong to the map $M_{c}$ and $M_{m}$ respectively. The alignment transformation matrix $T_{im\_ic}$ between the two keyframes is solved by the PNP algorithm ($M_{m}$ world coordinate system to $M_{c}$ world coordinate system transformation). We check $T^{Wm}_{ic\_im}\in SE(3)$ between $K_{c}$ and $K_{m}$ (the transformation matrix from $K_{m}$ to $K_{c}$ in the world coordinate system of $M_{m}$). If the yaw and 2-norm of the translation are both below the corresponding threshold, the assumption of position recognition will be accepted. If the location recognition assumption is accepted, the multi-map fusion module will be triggered. A short-term map fusion is carried out on the local window jointly defined by the keyframes around $K_{c}$ and $K_{m}$ and the map points observed by the keyframes, to optimize $T^{Wm}_{ic\_im}$. A long-term pose graph fusion optimization is performed for all the keyframes between the early position overlap keyframe and the nearest position overlap keyframe when the position overlap is accepted. The specific steps of the algorithm are as follows:

\begin{itemize}
	\item Local window. Multi-map fusion module respectively takes $M-1$ frame around the $K_{c}$ and $K_{m}$, and put them into $vKF_{c}$ and $vKF_{m}$ respectively, and puts $K_ {c}$ and $K_ {m}$ into the container correspondingly. Then, the 3D map point $MP_{m}$ observed in the keyframe of $vKF_{m}$ is projected onto the image coordinate system of $K_{c}$ according to the $T^{Wm}_{ic\_im}$ obtained previously. If the depth is negative or higher than the threshold, the matching of the point pair is rejected. Because of the same pose estimation accuracy, the larger the actual depth, the less reliable the point, and the worse the triangulation accuracy. The projection point in the pixel coordinate system is marked as $\rho$, then takes $\rho$ as the center of the circle and $r$ pixels as the radius to search feature points, and calculate the Hamming distance between $MP_{m}$ and the feature points. Then, the one with the smallest Hamming distance is selected as the candidate matching point. Then, the fundamental matrix is calculated by RANSAC to filter out the outer points, and the matching point pairs are respectively placed in $vMP_{c}$ and $vMP_{m}$. Then the local window makes a reverse projection to find more matching point pair. The keyframes in $vKF_{c}$ and $vKF_{m}$, and matching point pairs define a local window.
	
	\item Map fusion. As shown in Fig. \ref{fig.2.}, a local BA is executed to optimize $T_{im\_ic}$. If the optimized $T_{im\_ic}$ is accepted, the two submaps are fused into one map.
	
	\item Optimization of pose graph. As described in Fig. \ref{fig.3.}, if the $T_{im\_ic}$ solved in map fusion is accepted, global pose graph optimization between multiple submaps is performed. The drift corrected by the optimization results is propagated to subsequent map sequences.
\end{itemize}

\section{EXPERIMENTAL RESULTS}

For all the experiments presented in this section, the devices used are listed in Table \ref{tab.1.}, where PC is the machine used by VINS-Mono. To better evaluate the proposed system, we use the public dataset EuRoc\cite{2016The}. The datasets provide accurate ground truth data through Leica Total Station. Each sequence is processed by a separate agent, and the agent communicates with the server in real-time over the network. Taking into account the actual usage, the network has no special treatment, and the wireless campus network of the university is used.

\subsection{Precision evaluation of single agent mapping}
To evaluate the accuracy of our proposed system, we first evaluate the absolute trajectory error (ATE) of the keyframe trajectory. The Table \ref{tab.2.} compares the performance of VINS-Mobile1, VINS-Mobile2, VINS-Mobile2+Server, and VINS-Mono. VINS-Mobile1 adds the processing of reading text data, image flipping, and dedistortion on VINS-Mobile. We observe that the ATE of VIN-Mobile1 without loop optimization on MH\_02 is 0.048m. After the loop optimization is done, the error becomes 0.10m. The ATE without loop optimization on MH\_01 is 0.083m. After the loop optimization, it can reach 0.08m in the worst case. The loop optimization effect of VINS-Mobile1 on iphone6s is better than that on the iphone7p, even it has more computing resources. This is because the loop optimization overfitting the relative pose between the two matching frames, but this relative pose is subject to error. Based on the above analysis, we slightly modify the VINS-Mobile1 to VINS-Mobile2 by adding a filter processing which is used to avoid overfitting of loop optimization. Both VINS-Mobile1 and VINS-Mobile2 run a complete SLAM system on the mobile devices. VINS-Mobile2+Server runs VIO on the mobile devices and performs loop detection and global optimization on the server. VINS-Mono runs a complete SLAM system on the PC. Table \ref{tab.2.} shows that the VINS-Mobile2+Server mode has a slightly better effect than VINS-Mono except for the MH\_04 sequence. Experiment shows that the threshold value of the relative pose between two matching frames is too large, leading to the removal of the correct relative pose relationship. This threshold has not been tested much and may need to be adjusted further. Besides, the network we use is a little unstable because so many people are using it at the same time. If a better network is used, part of the experimental effect can be improved. Because during the experiment, we find that the same data is run 6 times, and the maximum error is 2cm different from the minimum error. This is because the relative pose between the two matching frames required for loop optimization is calculated in the sliding window. The delay of the network causes the matching frame to be removed from the sliding window, and the relative pose has not been calculated yet, resulting in this loop optimization not being performed. This is also the problem we need to solve later.

 \begin{table}[htbp]
	\caption{Hardware setup for experiments}
	\begin{center}
		\begin{tabular}{|c|c|c|}
			\hline
			Platform         & Type  &        Characteristics                         \\ \hline
			Server           & MacBook Pro      & i7 2.8GHz*16GB RAM          \\ \hline
			Agent 1          & iphone6s              & A9+M9 1.8GHz*2GB RAM       \\ \hline
			Agent 2          & iphone7p              & A10+M10 2.23GHz*3GB RAM      \\ \hline
			PC & Desktop computer & i7-9700k 3.60GHz*15.6GB RAM \\ \hline
		\end{tabular}
	\end{center}
	\label{tab.1.}
\end{table}
 
 \begin{figure}[htbp]
	\begin{center}
		\includegraphics[scale=0.20]{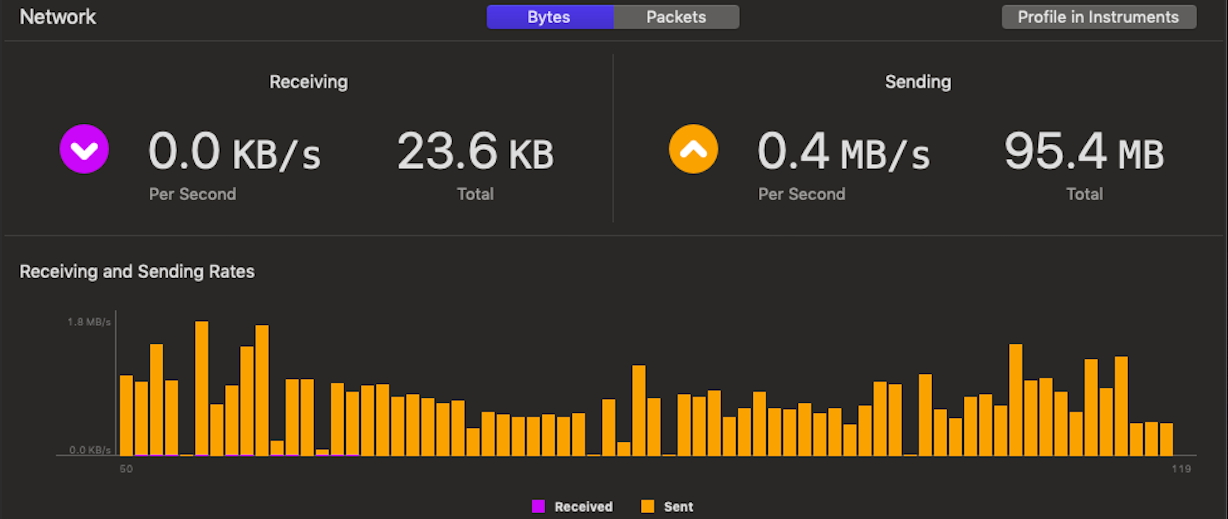}
	\end{center}
	\caption{Receiving on the left represents data information sent by the server to an agent, and sending on the right represents data information sent by an agent to the server. Among them, the per second module represents the real-time transmission rate, and the total module represents the amount of data transmitted during the operation of the entire system. In the table below, the main body is yellow, and only the bottom left row is purple. Each column represents the data transmission rate in that second, yellow represents the data rate sent, and purple represents the data rate received.}
	\label{fig.4.}
\end{figure}

\subsection{Bandwidth requirements}
When the agent is running , the source editor we used, i.e. the Xcode, can monitor the occupancy rate of the system resources. Fig. \ref{fig.4.} shows the real-time network occupancy rate of an agent, where the speed of receiving and sending data is real-time, not the average speed computed from total data volume divided by time, so the receiving data speed is 0.0KB/s, and similarly 0.4MB/s only represents the received speed of the last column in the bottom table in Fig. \ref{fig.4.}. The average bandwidth of an agent is about 0.8MB/s ($(95.4\text{Mb}+23.6\text{kb})/119\text{s}\approx0.8\text{MB/s}$). Besides, we can see that the transmitted data is relatively uniform, which also proves that, as described in the previous section 4.2, actively cutting large data packets and controlling the transmission frequency can effectively alleviate network congestion. Since the server has been fused and optimized in the later stage, the data fed back to the agent is more low-frequency.

 \begin{table*}[htbp]
	\caption{Comparison of absolute trajectory error between proxy local operation and proxy server mode}
	\begin{center}
		\begin{tabular}{|m{0.4cm}<{\centering}|m{1.2cm}<{\centering}|m{1.2cm}<{\centering}|m{1.2cm}<{\centering}|m{1.2cm}<{\centering}|m{1.5cm}<{\centering}|m{1.5cm}<{\centering}|m{2.5cm}<{\centering}|}
			\hline
			\multicolumn{1}{|c|}{\multirow{2}{*}{Dataset}} & \multicolumn{2}{c|}{VINS-Mobile1  ATE(m)} & \multicolumn{2}{c|}{VINS-Mobile2  ATE(m)} & \multicolumn{2}{c|}{VINS-Mobile2+Server  ATE(m)} & VINS-Mono  ATE(m) \\ \cline{2-8} 
			\multicolumn{1}{|c|}{}                         & iphone6s               & iphone7p              & iphone6s               & iphone7p              & iphone6s                  & iphone7p                  &   PC      \\ \hline
			MH\_01  & 0.076          & 0.070        & 0.061     & 0.061         & 0.053           & \textbf{0.050}          & 0.082    \\ \hline
			MH\_02  & 0.108      & 0.100    & 0.110     & 0.086     & 0.080           & \textbf{0.073}        & \textbf{0.073}     \\ \hline
			MH\_03  & 0.110        & 0.109     & 0.110       & 0.109        & \textbf{0.054}         & 0.066           & 0.057  \\ \hline
			MH\_04  & 0.167        & 0.170     & 0.167      & 0.166      & 0.166        & 0.163        & \textbf{0.148}  \\ \hline
			MH\_05  & \textbf{0.087 }     & 0.101     & 0.087     & 0.090         & 0.096        & 0.093         & 0.107 \\ \hline
		\end{tabular}
	\end{center}
	\label{tab.2.}
\end{table*}

\begin{table}[htbp]
	\caption{Map fusion absolute trajectory error comparison}
	\begin{center}
		\begin{tabular}{|c|c|c|c|c|c|}
			\hline
			\multicolumn{1}{|c|}{\multirow{2}{*}{Dataset}} & \multicolumn{5}{c|}{VINS-Mobile2+Server ATE(m)}                \\ \cline{2-6} 
			\multicolumn{1}{|c|}{}                         & 1st  & 2nd & 1st-2nd  & 2nd-1st & single agent(6s) \\ \hline
			MH\_01                                         & 0.119 & 0.101          & 0.071  & \textbf{0.046}  & 0.053        \\ \hline
			MH\_02                                         & 0.068   & 0.057          & 0.080  & \textbf{0.070}   & 0.080        \\ \hline
			MH\_03                                         & 0.068   & 0.144          & 0.067 & 0.102   & \textbf{0.054}      \\ \hline
			MH\_04                                         & 0.130  & 0.256          & 0.198  & 0.190  & \textbf{0.166}      \\ \hline
			MH\_05                                         & 0.173    & 0.119          & 0.159  & \textbf{0.090}   & 0.096      \\ \hline
		\end{tabular}
	\end{center}
	\label{tab.3.}
\end{table}

\begin{table}[htbp]
	\centering
	\caption{Multi-agent collaboration performance evaluation}
	\resizebox{\columnwidth}{!}{
		\begin{tabular}{|m{2.0cm}<{\centering}|m{1.5cm}<{\centering}|m{1.9cm}<{\centering}|m{1.8cm}<{\centering}|}
			\hline
			\multicolumn{1}{|c|}{\multirow{2}{*}{Dataset}} & \multicolumn{2}{c|}{VINS-Mobile2+Server ATE(m)} & VINS-Mono ATE(m)    \\ \cline{2-4} 
			\multicolumn{1}{|c|}{}                         & multi agent      & single agent(6s)      & multi-session \\ \hline
			MH\_03\&MH\_01                                       & \textbf{0.044}& 0.053                 & 0.047         \\ \hline
			MH\_01\&MH\_02                                       & \textbf{0.049}            & 0.080                  & \textbf{0.049}         \\ \hline
			MH\_04\&MH\_05                                       & \textbf{0.090}             & 0.095                 & 0.106         \\ \hline
			MH\_01\&MH\_05                                       & \textbf{0.092}            & 0.095                 & 0.117         \\ \hline
			MH\_02\&MH\_03                                       & 0.069            & \textbf{0.054}                 & 0.059         \\ \hline
			MH\_03\&MH\_04                                       & 0.127            & 0.166                 & \textbf{0.116}         \\ \hline
			MH\_05\&MH\_04                                       & 0.120             & 0.166                 & \textbf{0.096}         \\ \hline
			
	\end{tabular}}
	\label{tab.4.}
\end{table}

\subsection{Precision evaluation of multi-map fusion}
After analyzing the trajectory accuracy of a single agent, the accuracy of the fusion between multiple agents is evaluated. To better reflect the accuracy of the fusion, we divide each sequence of the EuRoc datasets into two parts, denoted as $A$ and $B$, and given to two agents respectively. The two agents run independently and only communicate with the server, and the server integrates the map information of the two agents. The second and third column in Table \ref{tab.3.} are the ATE of sequence $A$ and sequence $B$, respectively. The fourth column is the ATE of the global map obtained by merging sequence $B$ into the world coordinate system of the A sequence. The fifth column is the ATE of the global map obtained by merging sequence $A$ into the world coordinate system of sequence $B$. The sixth column is the ATE of a map obtained after an agent runs a complete data sequence and sends it to the server for loop optimization.

Table \ref{tab.3.} shows that the ATE of sequence $A$ and sequence $B$ is almost larger than that of a complete sequence. This is because the effect of map initialization has a greater impact on the ATE of the entire map, and small data sequences have less information. Moreover, $A$ and $B$ who are the main maps have a great influence on the accuracy of the final fusion. Because the proposed optimization strategy is to optimize the keyframes between the first matching frame and the last matching frame of the main map, while the other maps are optimized from the first frame of the map to the last matching frame. The drift of the optimization correction propagates to the subsequent map sequence, but the previous map sequence cannot be corrected unless the previous map sequence has a new loop constraint. Column 2 and column 3 in Table \ref{tab.3.} show that mapping only through VIO tracking is bad. It is found in the experiment that the filtering condition is too strong to ensure the accuracy of fusion, which leads to the common area detected in the previous sequence of the main map being regarded as not a correct hypothesis and cannot participate in the optimization of multi-map fusion.


 \subsection{ Multi-agent collaboration assessment}
The last experiment aims to analyze the ability of agents to share and reuse information in a collaborative environment. For this reason, we have experimented in single-agent mode, double-agent mode and VINS-Mono multi-session mode. Their ATE reported in Table \ref{tab.4.} are calculated by aligning the optimized global trajectory with the ground truth. For each experiment, in the case of multi-agent and VINS-Mono multi-session, we only align and evaluate the results of the second sequence. For the single-agent case, we only run the second sequence. Because there are very few overlapping regions in the MH\_02 and MH\_03, the proposed system only accepts two overlaps, resulting in the fusion error of the two submaps that can not be optimized. For the MH\_03 sequence, the ATE after the fusion optimization is higher than the ATE performing loop optimization only, which exists in the proposed system and VINS-Mono. The experiment shows that the other data sequences in Table \ref{tab.4.} have relatively more overlapping areas, and their mapping accuracy is higher than that in the single-agent mode. This shows that the proposed fusion optimization strategy is feasible, even if the relative pose of the two matched frames is inaccurate. But if there are fewer common areas between two submaps, the error will propagate to the subsequent map sequence and cannot be corrected. Improving the accuracy of the relative pose between two frames is an important problem we will solve in the future. Table \ref{tab.4.} also shows that the fusion accuracy of the proposed method has achieved similar results to VINS-Mono, and the benefits of sharing information obtained by multiple agents to improve accuracy during collaboration.

\section{Conclusions}
In this paper, we propose a cooperative monocular visual-inertial SLAM system for smart phones. The experiments show that the accuracy of mapping and fusion of the proposed SLAM system is comparable to VINS-Mono running on PC. Furthermore, our evaluation confirms that sharing information between participating agents during collaborative SLAM mapping can improve the accuracy of each agent pose estimation in real-time compared to a single agent scenario. To the best of our knowledge, the proposed system is the first visual inertial cooperative SLAM system running on the mobile phone and achieving two-way communication between the agent and the server. 


\balance

\bibliographystyle{IEEEtran}
\bibliography{ref}

\begin{thebibliography}{10}
\providecommand{\url}[1]{#1}
\csname url@samestyle\endcsname
\providecommand{\newblock}{\relax}
\providecommand{\bibinfo}[2]{#2}
\providecommand{\BIBentrySTDinterwordspacing}{\spaceskip=0pt\relax}
\providecommand{\BIBentryALTinterwordstretchfactor}{4}
\providecommand{\BIBentryALTinterwordspacing}{\spaceskip=\fontdimen2\font plus
\BIBentryALTinterwordstretchfactor\fontdimen3\font minus
  \fontdimen4\font\relax}
\providecommand{\BIBforeignlanguage}[2]{{%
\expandafter\ifx\csname l@#1\endcsname\relax
\typeout{** WARNING: IEEEtran.bst: No hyphenation pattern has been}%
\typeout{** loaded for the language `#1'. Using the pattern for}%
\typeout{** the default language instead.}%
\else
\language=\csname l@#1\endcsname
\fi
#2}}
\providecommand{\BIBdecl}{\relax}
\BIBdecl

\bibitem{qin2018vins}
T.~Qin, P.~Li, and S.~Shen, ``Vins-mono: A robust and versatile monocular
  visual-inertial state estimator,'' \emph{IEEE Transactions on Robotics},
  vol.~34, no.~4, pp. 1004--1020, 2018.

\bibitem{mourikis2007multi}
A.~I. Mourikis and S.~I. Roumeliotis, ``A multi-state constraint kalman filter
  for vision-aided inertial navigation,'' in \emph{Proceedings 2007 IEEE
  International Conference on Robotics and Automation}.\hskip 1em plus 0.5em
  minus 0.4em\relax IEEE, 2007, pp. 3565--3572.

\bibitem{bloesch2015robust}
M.~Bloesch, S.~Omari, M.~Hutter, and R.~Siegwart, ``Robust visual inertial
  odometry using a direct ekf-based approach,'' in \emph{2015 IEEE/RSJ
  international conference on intelligent robots and systems (IROS)}.\hskip 1em
  plus 0.5em minus 0.4em\relax IEEE, 2015, pp. 298--304.

\bibitem{weiss2012versatile}
S.~Weiss, M.~W. Achtelik, M.~Chli, and R.~Siegwart, ``Versatile distributed
  pose estimation and sensor self-calibration for an autonomous mav,'' in
  \emph{2012 IEEE International Conference on Robotics and Automation}.\hskip
  1em plus 0.5em minus 0.4em\relax IEEE, 2012, pp. 31--38.

\bibitem{lynen2013robust}
S.~Lynen, M.~W. Achtelik, S.~Weiss, M.~Chli, and R.~Siegwart, ``A robust and
  modular multi-sensor fusion approach applied to mav navigation,'' in
  \emph{2013 IEEE/RSJ international conference on intelligent robots and
  systems}.\hskip 1em plus 0.5em minus 0.4em\relax IEEE, 2013, pp. 3923--3929.

\bibitem{falquez2016inertial}
J.~M. Falquez, M.~Kasper, and G.~Sibley, ``Inertial aided dense \& semi-dense
  methods for robust direct visual odometry,'' in \emph{2016 IEEE/RSJ
  International Conference on Intelligent Robots and Systems (IROS)}.\hskip 1em
  plus 0.5em minus 0.4em\relax IEEE, 2016, pp. 3601--3607.

\bibitem{leutenegger2015keyframe}
S.~Leutenegger, S.~Lynen, M.~Bosse, R.~Siegwart, and P.~Furgale,
  ``Keyframe-based visual--inertial odometry using nonlinear optimization,''
  \emph{The International Journal of Robotics Research}, vol.~34, no.~3, pp.
  314--334, 2015.

\bibitem{sivic2003video}
J.~Sivic and A.~Zisserman, ``Video google: A text retrieval approach to object
  matching in videos,'' in \emph{null}.\hskip 1em plus 0.5em minus 0.4em\relax
  IEEE, 2003, p. 1470.

\bibitem{nister2006scalable}
D.~Nister and H.~Stewenius, ``Scalable recognition with a vocabulary tree,'' in
  \emph{2006 IEEE Computer Society Conference on Computer Vision and Pattern
  Recognition (CVPR'06)}, vol.~2.\hskip 1em plus 0.5em minus 0.4em\relax Ieee,
  2006, pp. 2161--2168.

\bibitem{galvez2012bags}
D.~G{\'a}lvez-L{\'o}pez and J.~D. Tardos, ``Bags of binary words for fast place
  recognition in image sequences,'' \emph{IEEE Transactions on Robotics},
  vol.~28, no.~5, pp. 1188--1197, 2012.

\bibitem{ye2017place}
Y.~Ye, T.~Cieslewski, A.~Loquercio, and D.~Scaramuzza, ``Place recognition in
  semi-dense maps: Geometric and learning-based approaches,'' 2017.

\bibitem{bonanni20173}
T.~M. Bonanni, B.~Della~Corte, and G.~Grisetti, ``3-d map merging on pose
  graphs,'' \emph{IEEE Robotics and Automation Letters}, vol.~2, no.~2, pp.
  1031--1038, 2017.

\bibitem{zhang2020map}
J.~Zhang, J.~Liu, K.~Chen, Z.~Pan, R.~Liu, Y.~Wang, T.~Yang, and S.~Chen, ``Map
  recovery and fusion for collaborative ar of multiple mobile devices,''
  \emph{IEEE Transactions on Industrial Informatics}, 2020.

\bibitem{campos2020orb}
C.~Campos, R.~Elvira, J.~J.~G. Rodr{\'\i}guez, J.~M. Montiel, and J.~D.
  Tard{\'o}s, ``Orb-slam3: An accurate open-source library for visual,
  visual-inertial and multi-map slam,'' \emph{arXiv preprint arXiv:2007.11898},
  2020.

\bibitem{elvira2019orbSLAM}
R.~Elvira, J.~D. Tard{\'o}s, and J.~Montiel, ``Orbslam-atlas: a robust and
  accurate multi-map system,'' \emph{arXiv preprint arXiv:1908.11585}, 2019.

\bibitem{schmuck2017multi}
P.~Schmuck and M.~Chli, ``Multi-uav collaborative monocular slam,'' in
  \emph{2017 IEEE International Conference on Robotics and Automation
  (ICRA)}.\hskip 1em plus 0.5em minus 0.4em\relax IEEE, 2017, pp. 3863--3870.

\bibitem{2018CCM}
------, ``Ccm-slam: Robust and efficient centralized collaborative monocular
  simultaneous localization and mapping for robotic teams,'' \emph{Journal of
  Field Robotics}, 2018.

\bibitem{2018CVI}
P.~Schmuck, M.~Karrer, and M.~Chli, ``Cvi-slam-collaborative visual-inertial
  slam,'' \emph{IEEE Robotics \& Automation Letters}, pp. 1--1, 2018.

\bibitem{cieslewski2018data}
T.~Cieslewski, S.~Choudhary, and D.~Scaramuzza, ``Data-efficient decentralized
  visual slam,'' in \emph{2018 IEEE International Conference on Robotics and
  Automation (ICRA)}.\hskip 1em plus 0.5em minus 0.4em\relax IEEE, 2018, pp.
  2466--2473.

\bibitem{cieslewski2017efficient}
T.~Cieslewski and D.~Scaramuzza, ``Efficient decentralized visual place
  recognition from full-image descriptors,'' in \emph{2017 International
  Symposium on Multi-Robot and Multi-Agent Systems (MRS)}.\hskip 1em plus 0.5em
  minus 0.4em\relax IEEE, 2017, pp. 78--82.

\bibitem{wang2019active}
W.~Wang, N.~Jadhav, P.~Vohs, N.~Hughes, M.~Mazumder, and S.~Gil, ``Active
  rendezvous for multi-robot pose graph optimization using sensing over
  wi-fi,'' \emph{arXiv preprint arXiv:1907.05538}, 2019.

\bibitem{li2017monocular}
P.~Li, T.~Qin, B.~Hu, F.~Zhu, and S.~Shen, ``Monocular visual-inertial state
  estimation for mobile augmented reality,'' in \emph{2017 IEEE International
  Symposium on Mixed and Augmented Reality (ISMAR)}.\hskip 1em plus 0.5em minus
  0.4em\relax IEEE, 2017, pp. 11--21.

\bibitem{2016The}
M.~Burri, J.~Nikolic, P.~Gohl, T.~Schneider, J.~Rehder, S.~Omari, M.~W.
  Achtelik, and R.~Siegwart, ``The euroc micro aerial vehicle datasets,''
  \emph{International Journal of Robotics Research}, vol.~35, no.~10, pp.
  1157--1163, 2016.

\end{thebibliography}

\end{document}